\documentclass[lettersize,journal]{IEEEtran}

\usepackage{amsmath,amsfonts,amssymb}

\usepackage{algorithmic}
\usepackage{algorithm}

\usepackage{array}
\usepackage{booktabs}
\usepackage{tabularx}
\usepackage{makecell}
\usepackage{multicol}
\usepackage{multirow}
\usepackage{threeparttable}
\usepackage{tablefootnote}

\usepackage[caption=false,font=normalsize,labelfont=sf,textfont=sf]{subfig}
\usepackage{graphicx}
\usepackage{stfloats}

\usepackage{siunitx}
\usepackage{gensymb}
\usepackage{textcomp}

\usepackage{xcolor}
\usepackage{soul}
\usepackage{url}
\usepackage{verbatim}

\usepackage[utf8]{inputenc}
\usepackage[english]{babel}
\usepackage{CJKutf8}

\usepackage{cite}

\definecolor{addedgreen}{RGB}{0,125,0}
\definecolor{revisionblue}{RGB}{0,90,220}

\soulregister{\cite}{7}
\soulregister{\citep}{7}
\soulregister{\citet}{7}
\soulregister{\ref}{7}
\soulregister{\pageref}{7}

\makeatletter
\@ifundefined{hl}{}{}
\makeatother

\begin{document}
\begin{CJK*}{UTF8}{gbsn}

\title{MCR-Bionic Hand: Anatomical Structural Priors for Dexterous Manipulation}

\author{
Haosen~Yang\textsuperscript{\dag},
%Yuyang Wei,
%Xiaotao Zhang,
%Lingyun Yan,
%Katie Mei,
%Pierre-Alexis Mouthuy,
%Darwin Caldwell,
and Guowu~Wei*,~\IEEEmembership{Member,~IEEE}

\thanks{\textsuperscript{\dag}Haosen Yang: ORCID 0000-0001-7488-6834.}

\thanks{Guowu Wei is with the School of Science, Engineering and Environment, University of Salford, Salford M5~4WT, UK.
(e-mail: g.wei@salford.ac.uk).}
\thanks{Corresponding author: Guowu Wei.}
}

\markboth{}
{Yang and Wei: MCR-Bionic: Anatomical Structural Priors for Dexterous Manipulation}

\maketitle

\begin{abstract}
Dexterous hand research often formulates grasping and in-hand manipulation as high dimensional active control problems governed by degrees of freedom, actuation capability, and control algorithms. Yet part of human hand dexterity is encoded in the physical architecture of bones, ligaments, tendons, aponeuroses, and intrinsic muscle pathways. This work describes this contribution as two linked forms of structural intelligence: structural prior generation, in which wrist--finger tenodesis, flexor digitorum superficialis/profundus routing, and the dorsal extensor hood transform low dimensional posture inputs into default grasp configurations and proximal and distal interphalangeal coordination; and muscle mediated modulation, in which extrinsic muscles, lumbricals, and interossei regulate metacarpophalangeal posture, distal stability, fingertip force paths, and contact states on top of the default state.

Based on this framework, MCR-Bionic Hand, hereafter MCR-Bionic, is developed as a 1:1 musculoskeletal biomimetic hand that integrates a two row eight bone wrist, cross wrist tendons, anatomical flexor routing, volar plate and collateral ligament constraints, the dorsal extensor hood, and intrinsic muscle pathways within one mechanical body. Functional demonstrations and geometric mechanical models show that wrist posture induces multi joint pre shaping, the extensor hood maps proximal interphalangeal posture to a coupled distal interphalangeal response, and intrinsic-plus pathways modulate distal stability and fingertip action direction after grasp formation. Contact rich tasks, including coin rotation, pen transfer, dorsal coin flipping, and cube manipulation, show that MCR-Bionic links low dimensional state generation with fine post contact modulation. These results suggest that anatomical biomimetics is valuable not for visual similarity or structural complexity, but for identifying human hand structures that perform part of control; once removed, what is lost is not morphology, but physical computation for dexterous manipulation.
\end{abstract}


% ===========================================================
\section{Introduction}
% ===========================================================

Whether robots need to be human like is not a question of appearance. For human like dexterous hands, the more important question is which structures are merely consequences of biological morphology, and which structures actually change the relations among input, motion, contact, and stability. If some anatomical structures preorganize grasping, coordinate distal joints, or stabilize contact states before active control intervenes, human likeness is not only a morphological goal, but may become a testable design principle for robotics.

Human hand dexterity is often attributed to high degrees of freedom, rich sensing, and complex neural control. However, part of this capability is not generated from scratch by a controller. The geometric topology of bones, ligaments, tendons, aponeuroses, and intrinsic muscle pathways redistributes moments, restricts unstable postures, and triggers multi joint coordination, giving the hand certain default operational tendencies before active control is applied\cite{valero2000predictive,garcia1991extensor}. Human like dexterity should therefore not be understood only as a high dimensional control problem, but also as a result jointly organized by structure and control.

This work focuses on three anatomical structural mappings directly related to hand operation. The first is wrist--finger tenodesis, in which cross wrist tendon paths convert changes in wrist posture into multi joint finger pre shaping\cite{wilson1956providing,shah2016evaluating,landsmeer1965mechanism}. The second is dorsal extensor hood differentiation, in which the geometric relations among the central slip, lateral bands, and terminal tendon allow the posture of the proximal interphalangeal joint (PIP) to influence the distal interphalangeal joint (DIP) and fingertip configuration\cite{craig1992anatomy,haines1951extensor,clavero2003extensor}. The third is the intrinsic-plus related pathway, in which the lumbricals and interossei connect metacarpophalangeal joint (MCP) posture, PIP/DIP stability, and fingertip contact state through the extensor hood\cite{li2000contribution,goin1980structural,hamada2015correction}. These structures are not equivalent to additional degrees of freedom. Their role is to form cross joint, posture dependent, and constrained sliding physical mappings. If default grasp posture, distal coordination, and part of the stability boundary are already embedded in hardware, control or learning no longer acts on an entirely unstructured high dimensional joint space, but on a physical system preorganized by anatomical topology.

Existing dexterous hands have made important progress in degrees of freedom, actuation integration, grasp types, and in-hand manipulation capabilities\cite{huang2025human,zhang2025biomimetic,zhang2025soft,zhao2025embedding,sankar2025natural,luo2025precise}. However, to achieve compact packaging, manufacturing reliability, and simplified control, many systems use ideal revolute joints, remote tendon routing, engineered linkages, or simplified wrist and extensor hood structures. These designs can produce effective grasping, but often return pre shape generation, distal coordination, and post grasp stability modulation to the control layer, although these functions could otherwise be supported by anatomical topology. The gap between robotic hands and human hands may therefore arise not only from limitations in actuators, sensing, and algorithms, but also from anatomical structures that are removed during engineering simplification even though they preorganize motion before control begins.

To test this hypothesis, this work proposes a top down engineering translation method: first identifying structural couplings in the human hand that contribute to manipulation, and then translating them into robot design rules that can be implemented, analyzed, and verified. Based on this method, this work presents MCR-Bionic, a musculoskeletal biomimetic hand platform. The platform integrates bones, ligaments, tendons, aponeuroses, and intrinsic muscle pathways directly related to grasping and in-hand manipulation within a single mechanical body, allowing wrist--finger coupling, extensor hood differentiation, and intrinsic-plus modulation to coexist. A single muscle single source closed loop hydraulic artificial muscle architecture further provides local in-hand activation for each muscle tendon pathway, allowing these structural mappings to be evaluated through task experiments and geometric mechanical models.

\textbf{The contributions of this work are as follows:}

1) A structural intelligence framework is formulated in which anatomical structures generate a default grasping state and muscle pathways modulate this state for dexterous manipulation.

2) MCR-Bionic is developed as a musculoskeletal robotic hand integrating wrist bones and ligaments, anatomical FDS/FDP routing, volar plate and collateral ligament constraints, the dorsal extensor hood, intrinsic muscle pathways, and local artificial muscle actuation.

3) Functional demonstrations and geometric mechanical models are used to show how wrist--finger tenodesis, extensor hood differential transmission, and intrinsic-plus modulation support grasp pre shaping, distal coordination, and in-hand manipulation.

Together, these contributions address the central scientific question of this work: whether robots need to be human like. The answer is not that robotic hands should replicate all biological details, but that a testable criterion is needed. When a human like structure changes the relations among input, motion, contact, and stability in the robotic body, and thereby generates default grasping, distal coordination, or post grasp modulation, human likeness becomes a functional principle in robotic design rather than a morphological target.

% ===========================================================
\section{Related work}
% ===========================================================

\subsection{Existing actuation methods}
In a hand, where the scale is small and the component density is high, direct joint actuation enables high bandwidth control through short torque paths and colocated sensing. However, it concentrates packaging volume and mass distally, provides limited intrinsic compliance, and often compromises output force\cite{kaneko2008development,takaki2010high}. Tendon driven designs that place power sources in the palm, wrist, or forearm reduce distal inertia and keep the hand compact, but long transmission paths introduce friction, backlash, posture dependent moment arm variation, and wrist interference, while also occupying arm volume\cite{junge2025adapt,singh2024validations}. Underactuated and differential mechanisms achieve robust enveloping grasps with fewer actuators, but usually sacrifice joint level controllability and in-hand degrees of freedom\cite{catalano2014adaptive,jin2024underactuated,abozaid2024soft}. Among fluidic actuators, pneumatic systems offer high compliance and high power density, but struggle to provide multichannel, high bandwidth, and precise control\cite{zhang2025soft,hua2025design,ou2024enhancing}. Centralized hydraulic systems provide higher stiffness and response, but channel coupling, valve complexity, and system noise increase substantially with scale\cite{malas2025fluidic,riener2023robots,wang2025bioinspired}. Material and mechanism based soft actuators can be miniaturized and impose low structural burden, but face limitations such as thermal hysteresis and lifetime in shape memory alloys, high voltage in dielectric elastomer actuators, environmental sensitivity in electroactive polymers, and nonlinearity and wear in twisted drives\cite{wang2024anthropomorphic,suthar2024design}.

\subsection{Existing robotic hands and the gap in anatomical replication}
Robotic hands for upper limb replacement and manipulation have evolved along three broad routes. Prosthetic hands emphasize comfort, low mass, and daily reliability, often using reduced degrees of freedom, underactuation, or compliant grasping to obtain robustness\cite{belter2013mechanical,gopura2017prosthetic}. Engineering grippers emphasize structural simplicity, efficient power routing, and high grasp success, often using differential mechanisms and compliant envelopes to adapt to diverse objects\cite{odhner2014compliant,deimel2016novel}. Research dexterous hands pursue high degrees of freedom, hybrid force and position control, and in-hand manipulation to test sensing, control, and learning algorithms\cite{santos2025shadow,grebenstein2012hand}. Each route has clear strengths, but most systems still rely on ideal revolute joints, remote tendon routing, and geometric guides or hard stops. As a result, they can more readily reproduce human like motions than the structural priors produced in the human hand by bones, ligaments, aponeuroses, and intrinsic muscles.

Recent musculoskeletal designs have begun to introduce bones, ligaments, and tendon routes into robotic hand prototypes\cite{xu2016highly,ozawa2014design,zhu2023partI}, suggesting that part of the control burden can be carried by structure. However, constraints in packaging, routing, and actuation integration mean that existing systems often still require remote actuation, simplified joints, or reduced wrist constraints. Cross wrist tendon paths, the extensor hood, and intrinsic muscle pathways therefore remain difficult to couple within a single platform. Robotics consequently lacks a full scale, anatomically homologous hardware platform for determining which dexterous capabilities arise from structural preorganization and which must be supplied by control or learning. Medicine also lacks a physical carrier that can map skeleton, ligament, and tendon parameters and simulate interventions such as ligament laxity, muscle imbalance, or tendon transfer\cite{sadati2022embodied,liu2023advances,sun2022digital}. MCR-Bionic is built as a structural intelligence research platform for this gap.

% ===========================================================
\section{Method}
% ===========================================================

\begin{figure*}[htb]
  \centering
  \includegraphics[width=1\textwidth]{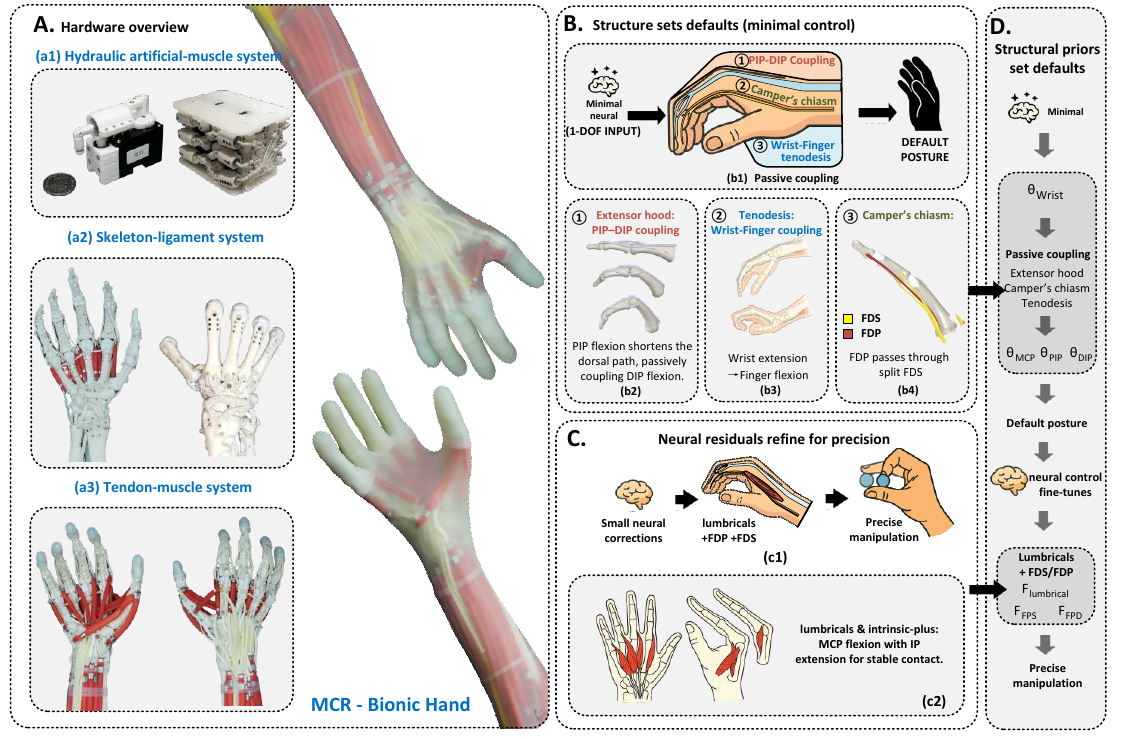}
  \caption{
  Implementation framework of anatomical pathways in MCR-Bionic.
  (A) The 1:1 musculoskeletal biomimetic hand platform, including the closed hydraulic artificial muscle system, skeletal and ligament system, and tendon and muscle system.
  (B) The structural prior layer consists of wrist--finger tenodesis, FDS/FDP tendon geometry, the extensor hood, and passive constraints.
  (C) The muscle modulation layer consists of active FDS/FDP inputs, extensor pathways, lumbricals, interossei, and intrinsic-plus related pathways.
  (D) The platform is used to examine how low dimensional structural input, extensor hood differentiation, and muscle modulation connect within the same physical system.
  }
  \label{fig06}
\end{figure*}

\subsection{Method framework}

The method consists of three steps. First, structural mappings directly related to grasping and in-hand manipulation are identified from human hand anatomy. Second, these mappings are translated into manufacturable, actuatable, and modelable paths of skeletons, ligaments, tendons, aponeuroses, and artificial muscles. Third, the resulting structures are examined within the musculoskeletal platform to determine whether they generate default grasping, distal coordination, and post grasp modulation.

This method follows human hand anatomy and reconstructs skeletons and soft tissues in engineering form to build a musculoskeletal biomimetic hand platform that can express structural mechanical intelligence (Fig.~\ref{fig06}A). The muscle system uses closed hydraulic artificial muscles: a motor driven transmission compresses or releases a sealed hydraulic chamber, causing fluid to enter or leave the hydraulic muscle and thereby produce contraction (Fig.~\ref{fig06}A1). This actuation architecture provides local in-hand activation for each muscle tendon pathway, allowing anatomical tendon routes, intrinsic muscle pathways, and structural couplings to coexist within the same hardware platform. The detailed mechanical configuration of the actuator and system level simulation are not analyzed in this work.

Fig.~\ref{fig06} shows the organization of these anatomical pathways in MCR-Bionic and serves as an implementation index for the structural prior layer and the muscle modulation layer described below.

\subsection{Structural prior layer: generation of default operational states (Fig.~\ref{fig06}B)}

The structural prior layer consists of geometric constraints, cross joint tendon routes, and passive limits that act before contact. Wrist--finger tenodesis serves as the proximal entry, transmitting wrist posture changes into whole finger pre shaping. The extensor hood serves as the distal entry, further distributing PIP posture to the DIP and fingertip configuration.

\textbf{1. Wrist--finger geometric coupling and cross wrist tendon length conservation}\cite{wilson1956providing,shah2016evaluating,landsmeer1965mechanism}:
As shown in Fig.~\ref{fig06}(b3), the flexor and extensor tendons cross the two row eight bone wrist and the extrinsic wrist ligaments/TFCC (triangular fibrocartilage complex). Small wrist motions change the tendon guiding angle and effective moment arm. During wrist extension, the cross wrist flexor path is lengthened and induces a tendency toward finger flexion. During wrist flexion, the opposite path is tensioned and promotes finger opening. This structure can be viewed as a passive feedforward link from proximal posture to distal opening and closing, making wrist angle a structural input for finger pre shaping.

Wrist--finger coupling is further shaped by the anatomical routing of the FDS and FDP. At the PIP, the FDS divides into two slips that insert into the middle phalanx and form the Camper chiasm, allowing the FDP to pass through and continue across the PIP and DIP to the distal phalanx (Fig.~\ref{fig06}(b4)). Before this split, the FDS runs more superficially and farther from the MCP rotation center, giving it a larger effective moment arm, whereas the FDP lies closer to the bone and has a smaller moment arm\cite{franko2011moment,yang2016assessing,kociolek2011modelling}. After the split, the two tendons approach a common plane at the PIP: FDS excursion is mainly absorbed by wrapping, whereas the sheath constrained FDP more closely follows a line angle relation. Thus, under equal proximal excursion, the FDS absorbs relatively more excursion at the MCP and less at the PIP, while the FDP shows the opposite tendency. This positional exchange creates an inherent excursion compensation mechanism around the PIP, limiting passive DIP flexion and allowing MCP/PIP flexion to form a thumb biased pinch preparatory posture.

In this layer, FDS/FDP participates in default configuration generation as cross wrist and interphalangeal tendon geometry. Its active muscle input is discussed in the subsequent modulation layer.

\textbf{2. Differential distribution of the extensor hood}\cite{craig1992anatomy,haines1951extensor,clavero2003extensor}:
As shown in Fig.~\ref{fig06}(b2), the extensor digitorum communis branches on the dorsal side of the proximal phalanx into the central slip and the left and right lateral bands. The central slip mainly acts on the PIP, whereas the bilateral lateral bands merge on the dorsal side of the middle phalanx and continue as the terminal tendon acting on the DIP. This topology forms a single input, multi segment differential network, allowing the extensor input to be redistributed among the MCP, PIP, and DIP according to joint posture and tendon geometry without requiring the control layer to specify the extension ratio of each joint independently.

Thus, the output of this layer is not a single joint angle, but a default operational state jointly determined by cross wrist tendon routing, the extensor hood, and passive joint limits.

\subsection{Muscle modulation layer: fine manipulation on the default state (Fig.~\ref{fig06}C)}

The muscle modulation layer acts on the default state described above. It does not redefine the whole hand posture, but regulates MCP posture, PIP/DIP stability, contact direction, contact force, and object posture on the basis of the existing pre shape, distal coordination, and contact relation.

\textbf{1. Endpoint force and motion regulation through extrinsic flexor and extensor pathways.}
On the default posture given by the structural prior layer, small inputs from FDS/FDP, extensor pathways, and intrinsic hand muscles, including the lumbricals and interossei, are used to close task residuals. FDS/FDP changes MCP/PIP/DIP flexion, endpoint force, and tangential motion through the flexor tendon paths. The extensor pathway participates in distal return, contact release, and fingertip direction regulation through the dorsal extensor hood network. This modulation layer mainly calibrates contact normal direction, contact force, and endpoint pose, thereby supporting fine operations such as pinching, rolling, and fingertip positioning.

\textbf{2. Intrinsic muscle pathways and angle dependent intrinsic-plus stabilization}\cite{li2000contribution,goin1980structural,hamada2015correction}:
As shown in Fig.~\ref{fig06}(c2), the lines of action of the lumbricals and interossei lie on the palmar side of the MCP rotation axis and can apply a flexion moment to the MCP. At the same time, these muscles enter the dorsal extensor hood and influence PIP/DIP extension stability through the lateral bands. With mild MCP flexion, intrinsic muscle input can simultaneously regulate MCP posture and stabilize extension of the IP joints, forming an intrinsic-plus posture. Because intrinsic muscle tension and extensor hood moment arms both vary with MCP angle, PIP/DIP stability and the ease of active flexion also become posture dependent.

\begin{figure*}[htb]
  \centering
  \includegraphics[width=1\textwidth]{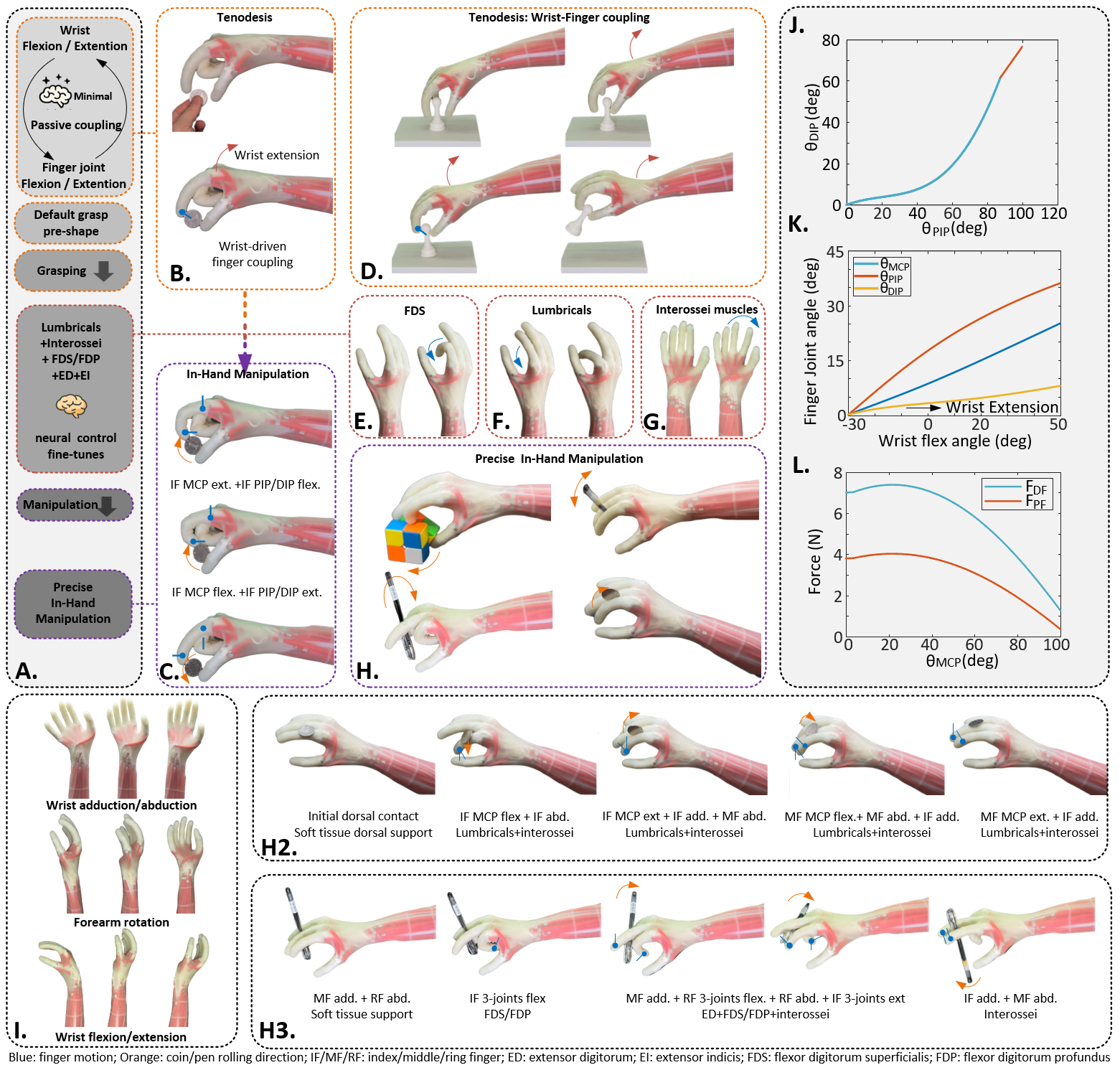}
\caption{
Functional demonstrations and geometric mechanical analysis of MCR-Bionic.
(A) Functional chain from wrist input to default grasping, muscle modulation, and fine in-hand manipulation.
(B,C) Wrist--finger tenodesis generates a default pinch and supports coin rotation.
(D) Wrist extension induces passive index finger flexion for chess piece grasping and lifting.
(E--G) Local modulation through FDS, lumbrical, and interosseous pathways.
(H) In-hand manipulation tasks, including Rubik's cube pushing, dorsal coin flipping(H2), pen transfer(H3), and pen swinging.
(I) Wrist range of motion.
(J--L) Models of the extensor hood, wrist--finger tenodesis, and intrinsic-plus pathways.
}
  \label{fig19}
\end{figure*}

% ===========================================================
\section{Results}
% ===========================================================

\begin{table*}[htbp]
  \centering
  \caption{{structural parameters and output specifications of the MCR-Bionic wrist and hand system.}}
  \label{tab:hand-all-params}
  \renewcommand{\arraystretch}{1.15}
  \setlength{\tabcolsep}{4pt}
  \begin{footnotesize}
  \begin{tabular}{llll}
    \toprule
    Parameter & Value & Parameter & Value \\
    \midrule

    \multicolumn{4}{l}{\textbf{Structural composition}} \\
    Number of bones, excluding the forearm
    & 23
    & Number of wrist ligaments
    & 61 \\

    Whole hand ligaments/soft tissue limit structures
    & $>103$
    & Number of muscle units
    & 46 \\

    Simplified hand DOF, excluding the wrist
    & 21
    & Wrist DOF
    & 3 \\

    Total simplified DOF, including the wrist
    & 24
    & DOF including small rotations/palmar deformation
    & $>45$ \\

    \midrule
    \multicolumn{4}{l}{\textbf{Long finger output performance}} \\
    MCP joint torque
    & 1.8 Nm
    & PIP joint torque
    & 1.2 Nm \\

    DIP joint torque
    & 0.5 Nm
    & Maximum fingertip force per finger
    & Approx. 20 N \\

    \midrule
    \multicolumn{4}{l}{\textbf{Thumb output performance}} \\
    Thumb MCP joint torque
    & 1.4 Nm
    & Thumb PIP joint torque
    & 0.8 Nm \\

    Thumb DIP joint torque
    & 0.5 Nm
    & Maximum thumb fingertip force
    & Approx. 20 N \\

    \midrule
    \multicolumn{4}{l}{\textbf{Wrist range of motion}} \\
    Wrist flexion/extension range
    & $-53^\circ$ to $18^\circ$
    & Wrist radial/ulnar deviation range
    & $-29^\circ$ to $19^\circ$ \\

    Wrist axial rotation range
    & $-60^\circ$ to $50^\circ$
    &
    & \\

    \bottomrule
  \end{tabular}

  \vspace{2pt}
  \raggedright
  \textit{Note:} Joint torques and fingertip forces are approximate values estimated from artificial muscle output forces and tendon moment arms.
  \end{footnotesize}
\end{table*}

\begin{table}[htbp]
\centering
\caption{Supplementary animations.}
\label{tab:supplementary-animations}
\renewcommand{\arraystretch}{1.05}
\setlength{\tabcolsep}{4pt}
\begin{footnotesize}
\begin{tabularx}{\columnwidth}{@{}l l X@{}}
\toprule
\textbf{File} & \textbf{Content} & \textbf{Purpose} \\
\midrule
S1 & PIP--DIP coupling & Distal coordination \\
S2 & Chess grasp & Default grasp generation \\
S3 & Coin rotation & Post grasp modulation \\
S4 & Dorsal coin transfer & Compliant contact manipulation \\
S5 & Pen transfer/swinging & Dynamic in-hand manipulation \\
S6 & Rubik's cube & Local modulation in grasp \\
\bottomrule
\end{tabularx}

\vspace{2pt}
\raggedright
\textit{Note: Supplementary videos are listed for reference and are not included in the preprint.}
\end{footnotesize}
\end{table}

MCR-Bionic is a musculoskeletal platform with anatomical level structural reconstruction. It integrates the skeletal and soft tissues summarized in Table~\ref{tab:hand-all-params}. Its joint limits and motion boundaries are implemented mainly through distributed compliant constraints formed by equivalent ligament, volar plate and tendon sheath.

The platform contains 23 bones, excluding ulna, radius and humerus, 61 wrist ligaments, more than 103 whole hand ligament/soft tissue, 46 muscle units, including 18 intrinsic hand muscles, and 24 simplified DOF with a 3-DOF wrist. When small rotations at the finger joints and palmar deformation are included, the movable DOF exceeds 45. The estimated maximum fingertip forces are approximately 20 N for the fingers and the thumb. The following experiments evaluate wrist driven grasp pre shaping, PIP--DIP distal coordination, and post contact muscle modulation on this platform.

The supplementary animations supporting these functional demonstrations are listed in Table~\ref{tab:supplementary-animations}.

\subsection{Structural priors generate default grasp configurations and distal coordination}

The first results examine two default operation components: whole finger pre shaping induced by wrist posture and PIP--DIP distal coordination mediated by the extensor hood.

To examine distal coordination, the extensor digitorum communis, central slip, lateral bands, and distal convergence path are reconstructed in the prototype (Fig.~\ref{fig19}(E)). PIP flexion and extension are transmitted to the DIP through changes in the lateral band path, causing the DIP to follow PIP posture without independent DIP actuation.

This PIP--DIP coupling is supported by two differential mechanisms. The first is constrained sliding of the lateral bands near the dorsal/lateral articular surface of the PIP, where joint posture changes the contact position, effective routing path, and equivalent length of the lateral bands, transmitting part of the proximal displacement to the DIP. The second is proximal differential compensation between the central slip and lateral band force transmission paths. This interface does not replace constrained sliding, but reproduces the elastic excursion distribution of the tendon aponeurosis network and provides an additional channel for relative displacement and length redistribution when friction, curvature discontinuity, assembly error, or wear weakens lateral band sliding. Thus, distal coordination arises from anatomically constrained sliding and elastic differential compensation; in this system, the extensor hood maps PIP posture to DIP and fingertip configuration through posture dependent distal distribution.

With distal coordination established, the chess grasping experiment examines wrist--finger coupling during default grasp generation (Fig.~\ref{fig19}(D)). In the initial state, the chess is placed on the table, the wrist is slightly flexed, the index finger and thumb are open without contacting the chess, and the index FDP and FDS remain tensioned without active linear pulling. Slight wrist extension alone changes the effective length of the cross wrist tendon paths, inducing the index finger and thumb to close toward the chess piece and pinch and lift it without additional active input to the finger joints.

This experiment shows that wrist posture can trigger the transition from no contact to stable thumb and index pinching under the tested condition. Together with the PIP--DIP result, it identifies proximal wrist--finger pre shaping and distal PIP--DIP coordination as two observable outputs of the reconstructed structural pathways.

\subsection{Muscle pathways transform default grasping into adjustable in-hand manipulation}

The next experiments examine post contact modulation of object posture, contact location, and fingertip action direction.

To examine this modulation entry, the lumbrical pathway is reconstructed so that it enters the dorsal extensor hood network from the palmar side and acts on the MCP and distal aponeurotic system (Fig.~\ref{fig19}(F)). Prototype results show that this pathway locally changes MCP posture without fully relying on whole finger pulling by extrinsic flexor and extensor muscles. Because its input spans the palmar and dorsal aponeuroses, MCP regulation also affects PIP/DIP distal stability, providing an entry for post grasp adjustment of fingertip force paths.

The coin rotation experiment shows the sequence from wrist driven contact formation to muscle mediated modulation (Fig.~\ref{fig19}(B),(C)). Initially, the remaining fingers are kept flexed to facilitate observation of index finger motion; the wrist is slightly flexed, the index FDP and FDS remain tensioned, and the coin is placed between the index fingertip and thumb fingertip without contact being established. Slight wrist extension alone then induces index finger flexion through wrist--finger coupling, allowing stable pinching of the coin. After the pinch is established, further FDP pulling with FDS relaxation rotates the coin toward the palm. Subsequent FDP relaxation, together with regulation through the dorsal extensor hood and lumbrical pathway, rotates the coin toward the fingertip.

The same hardware also performs multiple in-hand manipulation tasks (Fig.~\ref{fig19}(H)): dorsal coin transfer between the index and middle fingers(Fig.~\ref{fig19}(H2)), which also relies on interossei generated abduction/adduction (Fig.~\ref{fig19}(G)); pen transfer from the middle and ring finger space to the index and middle finger space through index finger motion(Fig.~\ref{fig19}(H3)); rapid pen swinging between the index and middle fingers; and Rubik's cube stabilization by the middle finger and thumb while the index finger pushes the cube surface.

Across these tasks, the hand maintains local contact while changing object posture, contact location, or tangential motion.

\subsection{Geometric models explain the mechanical origins of structural generation and muscle modulation}

To relate the prototype phenomena to specific geometric constraints, numerical analyses are performed for the extensor hood, wrist--finger tenodesis, and intrinsic-plus related pathways (Fig.~\ref{fig19}(J)--(L)). The full derivations are provided in Sec.~\ref{sec:mechanical intelligence-mechanisms}.

First, the PIP--DIP coupling model gives the distal joint angle relation mediated by the extensor hood (Fig.~\ref{fig19}(J)). The curve shows that the DIP does not linearly synchronize with the PIP, but exhibits nonlinear coupling with delayed onset followed by amplification: the DIP response is weak during early PIP flexion and increases with PIP flexion, with an increasing coupling slope.

Second, the wrist--finger coupling model shows that wrist angle changes are mapped through cross wrist tendon paths into multi joint MCP, PIP, and DIP responses (Fig.~\ref{fig19}(K)). The three joints do not change by equal amplitudes, but exhibit coordinated responses with different gains.

Finally, the intrinsic-plus model evaluates how MCP posture affects tension distribution in the intrinsic muscle extensor hood system (Fig.~\ref{fig19}(L)). As the MCP moves from extension into flexion, the dorsal fascicle tension $F_{\mathrm{DF}}$ and palmar fascicle tension $F_{\mathrm{PF}}$ both increase and then decrease. In mild MCP flexion, higher aponeurotic tension corresponds to stronger PIP/DIP extension stabilization and fingertip pad conformity. When the MCP flexes further into the enveloping grasp range, aponeurotic tension decreases and distal extension constraints weaken.

Together, the three models account for the observed distal differentiation, proximal pre shaping, and posture dependent stabilization in the prototype.

\begin{table*}[htbp]
\centering
\caption{Reference comparison of anatomical structural mappings in representative robotic hands and MCR-Bionic.}
\label{tab:anatomical-comparison}
\renewcommand{\arraystretch}{1.12}
\setlength{\tabcolsep}{2.0pt}
\begin{scriptsize}
\resizebox{\textwidth}{!}{
\begin{tabular}{p{2.35cm}cccccccccc}
\toprule
\textbf{System}
& \makecell{\textbf{Five finger}\\\textbf{full hand}}
& \makecell{\textbf{Active}\\\textbf{wrist}}
& \makecell{\textbf{Carpal bones--}\\\textbf{wrist ligaments}}
& \makecell{\textbf{Wrist--finger}\\\textbf{tenodesis}}
& \makecell{\textbf{FDS/FDP}\\\textbf{and tendon sheath}}
& \makecell{\textbf{Volar plate/}\\\textbf{collateral ligaments}}
& \makecell{\textbf{Dorsal}\\\textbf{extensor hood}}
& \makecell{\textbf{Intrinsic muscle}\\\textbf{pathways}}
& \makecell{\textbf{In-hand}\\\textbf{local actuation}}
& \makecell{\textbf{Post grasp}\\\textbf{structural modulation}} \\
\midrule

\multicolumn{11}{l}{\textit{{Conventional integrated dexterous hands}}} \\
Shadow Hand \cite{shadowhand_spec}
& Y & 2 DOF & N/A & N/A & Y & N/A & N/A & N/A & N/A & N/A \\

ILDA \cite{kim2021ilda}
& Y & N/A & N/A & N/A & N/A & N/A & N/A & N/A & Y & P \\

MCR-Hand III \cite{yang2021mcrhand}
& Y & 2 DOF & N/A & N/A & P & N/A & N/A & N/A & Y & P \\

\midrule
\multicolumn{11}{l}{\textit{{Anatomical/musculoskeletal biomimetic systems}}} \\
Allonic Hand \cite{allonic_hand}
& Y & N/A & N & N & Y & Y & Y & N & N & P \\

Clone Hand \cite{clone_hand}
& Y & Y & P & Y & Y & Y & Y & Y & Y & Y \\

ACT Hand \cite{deshpande2013act}
& P & 2 DOF & P & Y & P & P & P & P & N & P \\

ACB Hand \cite{tasi2019acb}
& Y & P & P & N & Y & Y & Y & Y & P & P \\

Xu--Todorov hand \cite{xu2016highly}
& Y & N/A & P & N & P & P & P & N & N & P \\

Single mat. hand \cite{tian2021single}
& P & N/A & P & N & Y & Y & P & N & N & P \\

BioFinger/hand \cite{zhu2023partI}
& P & N/A & Y & Y & Y & Y & Y & N & N & P \\

\textbf{MCR-Bionic}
& \textbf{Y}
& \textbf{3 DOF}
& \textbf{Y}
& \textbf{Y}
& \textbf{Y}
& \textbf{Y}
& \textbf{Y}
& \textbf{Y}
& \textbf{Y}
& \textbf{Y} \\

\bottomrule
\end{tabular}
}

\vspace{2pt}
\raggedright
\textit{Note:}
{Y, explicitly preserved; P, simplified or locally implemented; N, not implemented or not used as a design objective; N/A, not applicable to the reported system. Shadow Hand and ILDA are included as baseline systems. The table focuses on anatomical mappings that can affect motion, contact, and stability within a wrist and hand platform.}
\end{scriptsize}
\end{table*}

% ===========================================================
\section{Discussion}
% ===========================================================

\subsection{Structural priors change the starting point of control}

The results show that structural intelligence does not replace control, but changes the physical state on which control acts. In MCR-Bionic, part of the postural tendency, joint boundary, and force transmission path is preorganized by the skeleton, ligament, tendon, and aponeurosis topology. Control therefore acts on a physical system with default postures, stability boundaries, and preferred force paths.

The significance of this division is not limited to reducing degrees of freedom. The structural generation layer converts a small number of proximal inputs into multi joint pre shaping, allowing grasping to begin from a task relevant default configuration. The muscle modulation layer then changes distal stability, fingertip direction, and object posture after contact is established. Structure provides a constrained initial state, and active input further selects, adjusts, and uses that state.

The value of structural intelligence should therefore not be judged only by whether the number of actuators is reduced. A more relevant criterion is whether structural mappings change the initial conditions, stability boundaries, and reachable paths of the control problem. A high DOF system without structural priors may still need the control layer to compensate for posture generation, joint coordination, and contact stability. A system with appropriate structural mappings shifts part of these functions into the body, allowing control to focus more on task specific modulation.

\subsection{Wrist--finger coupling and the extensor hood form a precontact grasping chain}

Wrist--finger coupling and the extensor hood provide structural mappings at different scales during default grasping. The former introduces wrist posture changes into the whole finger system, so that wrist motion changes not only palm pose, but also the effective length of cross wrist tendon paths and the internal tension state of the fingers. The latter redistributes displacement and tension within the distal extensor apparatus, allowing PIP posture to influence the DIP and fingertip configuration.

When functionally connected, these two mappings make grasp pre shaping different from a set of joint by joint commands. A small proximal active input or wrist posture adjustment can form multi joint pre shaping through cross wrist tendon paths. The extensor hood then maps PIP motion into DIP distal coordination, whereas the volar plates and collateral ligaments define the stability boundaries within which these synergies occur. Unlike compliant mechanisms that mainly rely on passive deformation after object contact, this chain changes finger configuration before contact occurs, forming a precontact structural prior triggered by low dimensional control.

The design value of wrist--finger coupling and the extensor hood is therefore not to generate a fixed grasp posture, nor to eliminate control. It is to transform a small amount of control input into a default mechanical state that subsequent muscle modulation can continue to use. Proximal structural input provides whole finger pre shaping, and distal extensor hood differentiation constrains DIP and fingertip posture, allowing grasping to start from a physical state that already contains coordination and boundaries.

\subsection{Intrinsic-plus related pathways support post grasp contact modulation}

Intrinsic-plus related pathways act after the default grasp has formed. Their main function is not to create the initial contact, but to change force direction, contact points, and distal posture while contact is maintained. For in-hand manipulation, the main difficulty often lies not in establishing contact, but in satisfying contact maintenance and object regulation at the same time.

The coin rotation experiment illustrates this role. Wrist--finger coupling first forms a default pinch between the index finger and thumb. When the coin needs to rotate back from the palmar direction toward the fingertip direction, the system cannot rely only on extensor hood pulling to extend the PIP/DIP, because the pinch force would become insufficient to maintain object contact. At this stage, lumbrical contraction and extensor hood tensioning act together, allowing MCP posture regulation and PIP/DIP distal stability to occur simultaneously. The reverse rotation of the coin therefore corresponds to force path redistribution under maintained contact, rather than simple finger extension.

The value of the lumbricals and interossei is therefore not to add another independently controlled degree of freedom, but to provide a functional path from MCP posture to extensor hood tension, and then to distal stability and fingertip contact state. A robotic hand without this pathway may still enter a grasping state through wrist--finger coupling, but it is more difficult to transform that state into directionally controlled fine manipulation.

\subsection{Boundary of anatomical reconstruction: functional fidelity rather than morphological copying}

The results refine the question of whether robots need to be human like. Human like structure is valuable not because it increases morphological similarity, degrees of freedom, or structural complexity, but because it preserves functional mappings that change the relations among input, motion, contact, and stability.

Table~\ref{tab:anatomical-comparison} places this criterion in the context of representative robotic hands. Shadow Hand, ILDA and MCR-Hand III serve as baselines for highly integrated dexterous hands, whereas the remaining systems represent anatomical or musculoskeletal biomimetic approaches. The table does not rank performance; it identifies which task relevant structural mappings are physically preserved within a single wrist and hand platform. Existing systems have advanced commercial maturity, actuation integration, degrees of freedom, task capability, or local anatomical mechanisms. MCR-Bionic differs in integrating skeletons, the wrist ligament network, volar plates and collateral ligaments, tendon sheaths, FDS/FDP routing, the dorsal extensor hood, intrinsic muscle pathways, and local artificial muscle actuation in the same wrist and hand system, allowing these structures to act as continuous hardware priors.

Functional fidelity also depends on how motion is bounded. In the human hand, joint limits are not imposed by ideal revolute axes and rigid stops alone, but by distributed compliant constraints formed by ligaments, volar plates, tendon sheaths, joint capsules, and surrounding soft tissues. A robotic hand may reproduce similar joint angle ranges with rigid stops, yet lose micro off axis motion, passive adaptation, and tension redistribution under contact. MCR-Bionic preserves such compliant boundaries through equivalent soft tissue limits, allowing the fingers to adjust contact points, fingertip direction, and distal posture during contact rich tasks such as coin flipping, dorsal transfer, and pen manipulation.

This is why the platform adopts a relatively complete anatomical reconstruction. Completeness is not the objective itself; it reduces the risk of removing functional mappings before they are understood. In the human hand, bones, ligaments, tendons, the extensor hood, intrinsic muscles, and soft tissue limits act through relative position, cross joint routing, passive constraint, compliant boundary formation, and tension transmission. Removing a structure too early may remove not a morphological detail, but a physical relation that exists before control begins.

For human like dexterous hands, the design question is therefore not how many anatomical details should be copied, but which mappings and compliant constraints must be preserved. When a structure generates default grasping, maintains distal coordination, provides compliant motion boundaries, or supports fine post contact modulation, it becomes a functional structural prior rather than a biomimetic appearance feature.

\subsection{Validation scope and future quantification}

The validation strategy must match the physical nature of MCR-Bionic. The platform reconstructs a continuous musculoskeletal system formed by irregular bone articular surfaces, ligaments, tendons, and muscle pathways, rather than an ideal linkage mechanism with revolute joints and rigid links. This work therefore does not attempt a full analytical model of every soft tissue contact, frictional interaction, pretension state, or small deformation. Instead, the analysis focuses on three observable functional mappings and cross validates them with prototype behavior and task performance.

Destructive controls are also not used. In a highly coupled musculoskeletal system, removing a ligament, an extensor hood band, or an intrinsic muscle pathway simultaneously changes joint boundaries, tendon path length, pretension, friction, and contact compliance, making single structure attribution difficult. Repeatedly fabricating equivalent control prototypes that differ only in one target function is also impractical because the prototype depends on complex manual assembly and soft tissue tuning. Destructive durability tests would consume the current platform without directly addressing the structural mapping mechanisms studied here.

Future validation should therefore emphasize measurable and repeatable quantification while preserving the complete anatomical pathways. Tendon tension, joint angle, fingertip contact force, and object motion tracking can establish a quantitative loop from wrist input, tendon excursion, and extensor hood tension to contact state. Adjustable pretension, replaceable extensor hood materials, simplified tendon routes, and control compensation schemes can then compare the actuation input, feedback information, and control complexity required for the same grasp pre shaping, distal coordination, and in-hand manipulation.

% ===========================================================
\section{Anatomical reconstruction of the musculoskeletal robotic hand}
% ===========================================================

\begin{figure*}[htb]
  \centering
  \includegraphics[width=\textwidth]{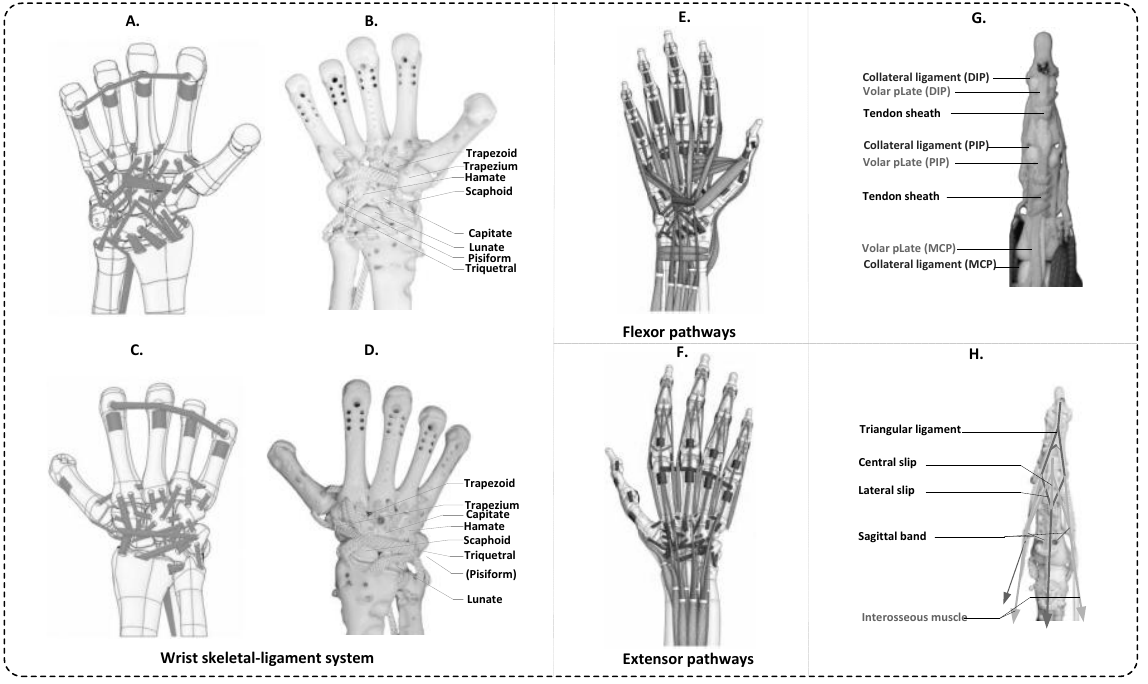}
\caption{Anatomical reconstruction of MCR-Bionic.
(A--D) Wrist skeletal and ligament system and corresponding anterior/posterior prototype views.
(E,F) Anterior and posterior views of the tendon muscle architecture.
(G) Palmar finger reconstruction and prototype, showing MCP/PIP/DIP volar plates, collateral ligaments, and tendon sheaths.
(H) Dorsal finger reconstruction and prototype, showing the extensor hood, sagittal band, and extensor tendon routes.}
\label{fig05}
\end{figure*}

\subsection{Design of the biomimetic robotic hand}

\vspace{6pt}
\noindent\textbf{1.\quad Wrist bones and ligamentous architecture}
\vspace{6pt}

As shown in Figs.~\ref{fig05}(A)--(D), the wrist reconstructs a two row, eight bone carpal architecture with ligamentous constraints on the palmar and dorsal sides. The proximal row contains the scaphoid, lunate, triquetrum, and pisiform, whereas the distal row contains the trapezium, trapezoid, capitate, and hamate. These bones are arranged into radial, central, and ulnar functional columns and are constrained by extrinsic and intrinsic wrist ligaments, including palmar, dorsal, intercarpal, collateral, and carpometacarpal connections. The resulting structure is not a simple revolute wrist, but a compact multi bone, multi ligament assembly that provides compliant wrist boundaries and the proximal anatomical basis for wrist driven tendon coupling.

\vspace{6pt}
\noindent\textbf{2. Finger skeleton, ligaments, and extensor hood}
\vspace{6pt}

As shown in Figs.~\ref{fig05}(G) and (H), each finger reconstructs a serial MCP/PIP/DIP skeletal chain together with palmar joint limits, collateral stabilization, tendon sheaths, and a dorsal extensor apparatus. The volar plates and collateral ligaments define compliant motion limits at the MCP, PIP, and DIP joints, while the tendon sheaths guide flexor and extensor tendon routing around small joint spaces. On the dorsal side, the extensor tendon divides into the central slip and lateral bands, which continue through the terminal tendon and are stabilized by the sagittal band and associated hood structures. This layered reconstruction is technically demanding because joint limits, tendon guidance, and extensor hood transmission must occupy the same narrow anatomical space while preserving both flexibility and alignment.

The intrinsic muscles enter this dorsal apparatus through the lumbrical and interosseous pathways. Their lines of action pass on the palmar side of the MCP axis and then couple into the extensor hood, allowing MCP flexion, PIP/DIP extension stabilization, and finger abduction/adduction to coexist within the same tendon network.

\vspace{6pt}
\noindent\textbf{3. Tendon muscle architecture}
\vspace{6pt}

Figs.~\ref{fig05}(E) and (F) summarize the anterior and posterior tendon muscle architecture. The dorsal side integrates wrist extensors, finger extensors, thumb extensors/abductors, and their entry into the extensor hood. The palmar side integrates wrist flexors, FDS/FDP flexor pathways, thenar and hypothenar muscles, and the intrinsic muscle routes. Together, these pathways form a dense in hand actuation architecture in which extrinsic flexor/extensor routes and intrinsic muscle routes are co-located with the reconstructed skeletal, ligamentous, and soft tissue boundaries.

This design preserves the main anatomical organization needed for wrist motion, finger flexion/extension, thumb motion, tendon excursion, extensor hood transmission, and intrinsic muscle modulation. Its difficulty lies not in reproducing isolated parts, but in integrating bones, ligamentous limits, tendon sheaths, extensor hood structures, intrinsic pathways, and local artificial muscle units within a 1:1 hand scale platform without eliminating the compliant interactions among them.

Implementation-level details, including detailed CAD geometry, anchoring layouts, material parameters, pretension settings, assembly procedures, and higher resolution structural figures, are outside the scope of this preprint and will be released with subsequent documentation.

% ===========================================================
\section{Mechanistic explanation of mechanical intelligence in the robotic hand}
\label{sec:mechanical intelligence-mechanisms}
% ===========================================================

\subsection{PIP--DIP structural coupling mechanism based on the extensor hood}

\begin{figure*}[htb]
  \centering
  \includegraphics[width=\textwidth]{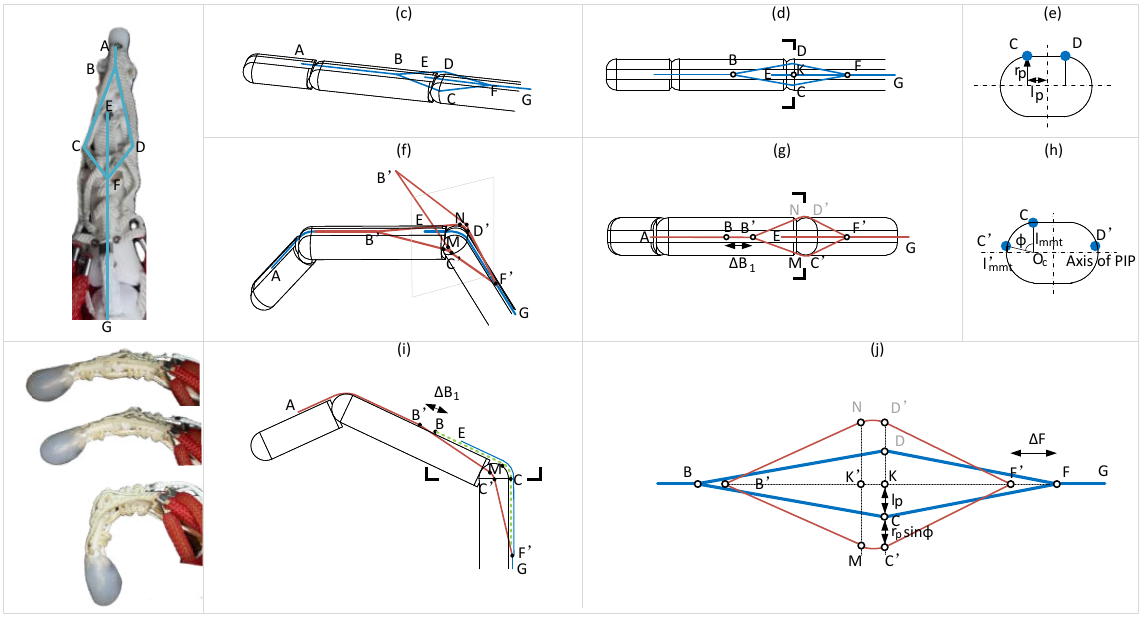}
\caption{Working mechanism of the extensor hood.
(a,b) Prototype views showing the central slip, lateral bands, and PIP/DIP flexion.
(c--j) Schematics of the extensor hood coupling: (c--e) extended state in lateral, top, and PIP cross sectional views; (f--h) coupled PIP/DIP flexion in the corresponding views; (i) lateral configurations at different flexion levels; (j) geometric comparison between the extended state (blue) and flexed state (red).}
\label{fig02}
\end{figure*}

In the extensor hood network, the proximal input enters at $G$ and divides into three branches at $F$ on the dorsal side of the proximal phalanx (Fig.~\ref{fig02}(a)). The central slip $FE$ crosses the PIP and inserts at $E$ on the dorsal side of the middle phalanx. The two symmetric lateral bands $FCB$ and $FDB$ contact the dorsal/lateral PIP arc surface at $C$ and $D$, converge at the gliding junction $B$ on the dorsal side of the middle phalanx, and connect to the distal point $A$ through the terminal tendon. Here, $B$ is a gliding convergence point rather than a fixed insertion. In full PIP/DIP extension (Fig.~\ref{fig02}(c),(d)), the central slip maintains PIP extension, whereas the bilateral lateral bands preserve tension continuity along the dorsal $C/D$--$B$--$A$ paths. The dorsal/lateral PIP contact geometry is approximated as a semi capsular arc surface (Fig.~\ref{fig02}(e)). Under the dorsal retaining effect of the extensor hood and its supporting structures, $C$ and $D$ remain on the dorsal or dorsolateral side of the PIP rotation axis and do not cross to the palmar side.

During coupled PIP/DIP flexion (Fig.~\ref{fig02}(f),(g); lateral view in Fig.~\ref{fig02}(i)), flexor pulling changes the proximal input, and the bifurcation point $F$ is released toward the PIP and moves to $F'$. As the PIP flexes, the tendon arc contact points $C,D$ undergo constrained sliding on the semi capsular arc surface to $C',D'$, and the lateral band paths become $F'C'B'$ and $F'D'B'$. The cross section in Fig.~\ref{fig02}(h) shows that the effective moment arm of tendon $F'C'B'$ about the PIP rotation axis decreases from $l_{mmt}$ to $l'_{mmt}$. This moment arm reduction and the contact point migration $C,D\to C',D'$ describe the same geometric process: the forward motion of $F'$ drives constrained sliding along the dorsal/lateral arc, shortening the actual lateral band wrapping path over the dorsal PIP surface.

Fig.~\ref{fig02}(i) compares the virtual path without contact point sliding (green dashed line) with the actual path after constrained sliding (red solid line). The actual cross joint path is shorter when passing over the dorsal PIP arc. This shortening is defined, relative to the no sliding case, as $\Delta B_1$. On the dorsal side of the middle phalanx, the shortened path is equivalent to a distal release of the gliding convergence point $B'$ by $\Delta B_1$ along the terminal tendon direction, which is transmitted distally along the $B'\!-\!A$ path. This release does not mean that the dorsal terminal tendon actively pulls the DIP into flexion. Rather, in the presence of flexor input, it reduces the terminal tendon extension constraint on the DIP and provides the tendon length allowance required for coupled DIP flexion.

From the top view (Fig.~\ref{fig02}(d),(g), compared with Fig.~\ref{fig02}(i)), the sliding of $C',D'$ toward the lateral sides of the semi capsular arc decreases the lateral opening angle of the quadrilateral $B'$--$C'$--$F'$--$D'$ and mildly retracts $B'$ toward the PIP, away from the DIP. This lateral effect counteracts distal release and is denoted by $\Delta B_2$, with a direction opposite to $\Delta B_1$. Thus, during flexion, the net migration of $B'$ is governed mainly by two geometric effects: cross arc path shortening $\Delta B_1$ caused by constrained sliding, and retraction $\Delta B_2$ caused by lateral spreading in the top view. If dorsal retention is maintained and the $\Delta B_1$ effect dominates, PIP flexion produces distal release at $B'/A$ through lateral band path redistribution and indirectly induces coupled DIP flexion.

If the lateral bands do not slide against the dorsal/lateral PIP arc surface and instead behave as uncoupled single cross joint paths, the actual tendon path across the PIP does not effectively shorten during flexion, and the distal release source represented by $\Delta B_1$ is absent. Proximal input is then mainly consumed at the PIP, making an effective coupled DIP response difficult. Dorsal retention, constrained sliding, and a lateral band path that does not cross the joint axis are therefore the geometric conditions for this PIP--DIP structural coupling.

The process reverses during extension. When the extensor tendon is pulled proximally, $F'$ retreats to $F$, the PIP extends, and the contact points $C',D'$ slide back along the arc surface to $C,D$. The PIP effective moment arm increases from $l'_{mmt}$ to $l_{mmt}$, strengthening the extension moment arm. Meanwhile, the lateral band path across the dorsal PIP arc changes from short to long, corresponding to a sign reversal of the flexion induced shortening $\Delta B_1$. The system recovers tendon length from the distal side, so that $B'$ retracts proximally and retensions the dorsal DIP path through the terminal tendon. The top view lateral geometry partially counteracts this recovery through $\Delta B_2$, but under dorsal retention and constrained sliding, the main result remains: PIP extension recovers tendon length and drives proportional return of the DIP with the PIP.

To quantify this PIP--DIP structural coupling, a reduced order geometric model based on tendon length conservation is established. The model does not describe continuous deformation of the extensor hood material, but retains the geometric constraints directly related to the coupled motion: approximately inextensible lateral bands, constrained sliding on the dorsal/lateral PIP arc surface, and length transfer between the gliding convergence point $B$ and the terminal tendon. Unless otherwise stated, flexion angles are positive, and distal migration of $B$ along the terminal tendon direction is defined as positive release. Equalities below denote geometric relations within the equivalent model; real three dimensional extensor hood deformation and local contact errors are absorbed by higher order compensation terms and by the range of model validity.

\paragraph{Model variables and range of validity}
Considering proximal input with flexor participation, the bifurcation point $F$ undergoes an axial displacement $\Delta F$ along the finger axis. The geometric states of the extensor hood during PIP/DIP extension and flexion are shown in Fig.~\ref{fig02}(j). The dorsal/lateral PIP contact geometry is approximated as a semi capsular arc surface (Fig.~\ref{fig02}(h)), with equivalent radius $r_p$, equivalent center $O_c$, and cross sectional eccentricity $l_p$. The PIP and DIP flexion angles are denoted by $\theta_{\text{PIP}}$ and $\theta_{\text{DIP}}$, respectively, and the sweep angle of the representative contact point $C$ relative to $O_c$ is denoted by $\phi$. During constrained sliding, a linear sliding parameterization is used:
\begin{equation}
\phi = k\,\theta_{\text{PIP}},
\qquad
0 \leq \phi \leq \phi_{\max} < \frac{\pi}{2},
\label{eq:phi_k_theta}
\end{equation}
where $k$ is the geometric proportionality coefficient between contact point sliding and PIP flexion, and $\phi_{\max}$ is the sliding limit of the contact point on the dorsal/lateral arc surface. The closed form model applies to the constrained sliding stage,
\begin{equation}
0 \leq \theta_{\text{PIP}} \leq \theta_{\text{PIP},\lim},
\qquad
\theta_{\text{PIP},\lim}=\frac{\phi_{\max}}{k}.
\label{eq:theta_limit}
\end{equation}

The effective wrapping radius of the terminal tendon at the DIP is denoted by $r_d$ and is obtained from the prototype or CAD geometry. The quantities $l_{FK}$, $l_{FCB}$, and $l_{BK}$ denote the geometric lengths of the $F\!-\!K$, $F\!-\!C\!-\!B$, and $B\!-\!K$ segments in the reference configuration (Fig.~\ref{fig02}(f),(g),(j)). These lengths are measured from the prototype or CAD model. The parameters $k$ and $k_B$ describe the contact point sliding ratio and higher order path compensation, respectively.

\paragraph{Proximal input and PIP flexion angle}
In the equivalent model, if the proximal input is mainly absorbed by the dorsal/lateral PIP arc, the axial release of the bifurcation point $F$ is parameterized by the arc length relation
\begin{equation}
\Delta F = r_p\,\theta_{\text{PIP}} .
\label{eq:dF_from_PIP}
\end{equation}
This equation is an input parameterization within the model and does not imply that all proximal displacement in the real structure is strictly converted into PIP rotation. Combining Eq.~\eqref{eq:phi_k_theta} with Eq.~\eqref{eq:dF_from_PIP} gives
\begin{equation}
\phi = k\,\frac{\Delta F}{r_p}.
\label{eq:phi_from_dF}
\end{equation}

\paragraph{Dominant distal release generated by constrained sliding}
When the contact point $C$ slides along the semi capsular arc surface to $C'$, the actual tendon path on the dorsal side of the PIP shortens relative to the no sliding virtual path. In the cross sectional equivalent model, contact point sliding projects the effective moment arm of the tendon segment crossing the PIP arc from $r_p$ to $r_p\cos\phi$. The dominant path shortening amount is therefore written as
\begin{equation}
\Delta B_1 =
\bigl(r_p-r_p\cos\phi\bigr)\,\theta_{\text{PIP}} .
\label{eq:DeltaB1}
\end{equation}
This term captures the combined effect of local path shortening and reduced effective moment arm. The required wrapping length across the dorsal PIP arc decreases with constrained sliding and provides the main source of distal release of $B'$ along the terminal tendon direction.

\paragraph{Retraction compensation caused by lateral spreading}
During flexion, the constrained sliding $C\!\to\!C'$ and $D\!\to\!D'$ changes the sagittal wrapping path of the lateral bands and produces lateral spreading of the quadrilateral $B'\!-\!C'\!-\!F'\!-\!D'$ in the top view. This spreading retracts $B'$ toward the PIP. To quantify this effect, the contact between $F'C'B'$ and the arc surface is treated as a short arc $C'M$ (Fig.~\ref{fig02}(f),(j)), with length
\begin{equation}
l_{C'M} = r_p\cos\phi\,\theta_{\text{PIP}} .
\label{eq:LCM}
\end{equation}

To measure the retraction in a common plane, $B'MC'$ and its mirrored structure $B'ND'$ are rigidly unfolded into the plane containing $F'C'D'$, preserving all segment lengths (Fig.~\ref{fig02}(f),(j)). The proximal tendon length $F'C'$ is then given by planar geometry:
\begin{equation}
l_{F'C'} =
\sqrt{
\bigl(l_{FK}-\Delta F\bigr)^2
+
\bigl(l_p+r_p\sin\phi\bigr)^2
}.
\label{eq:LFCp}
\end{equation}
Tendon length conservation of the same lateral band before and after flexion gives
\begin{equation}
l_{B'M} =
l_{FCB}-l_{C'M}-l_{F'C'} ,
\label{eq:inext}
\end{equation}
where $l_{B'M}$ is the length of segment $B'M$, and $l_{FCB}$ is the total length of the continuous path $F\!-\!C\!-\!B$ in the reference configuration. Projection of $B'M$ along the midline direction further gives
\begin{equation}
l_{B'K'} =
\sqrt{
l_{B'M}^{2}
-
\bigl(l_p+r_p\sin\phi\bigr)^2
},
\qquad
l_{B'K} \approx l_{B'K'}+l_{C'M}.
\label{eq:BKprime}
\end{equation}
The expression for $l_{B'K}$ includes planar unfolding and path splicing approximations, and is therefore written with approximate equality. Using the measured constant $l_{BK}$ in the reference configuration as the baseline, the retraction compensation caused by lateral spreading is defined as
\begin{equation}
\Delta B_2 = l_{BK}-l_{B'K}.
\label{eq:DeltaB2}
\end{equation}
Under the positive direction convention, if lateral spreading moves $B'$ closer to the PIP, then $l_{B'K}<l_{BK}$, so $\Delta B_2>0$ and enters Eq.~\eqref{eq:DeltaB_net} as a term that counteracts distal release.

\paragraph{Local wrapping and second order geometric compensation}
In addition to constrained sliding and lateral spreading, local wrapping and second order geometric effects occur during flexion. As shown in Fig.~\ref{fig02}(i), during $C\!\to\!C'$, elongation of $F'C$ into $F'C'$ increases the wrapping length on the proximal phalanx side, and slight deflection of $B'M$ also affects the position of $B'$ at second order. As $\theta_{\text{PIP}}$ increases, the tendon near the $F'$ end of $F'C'$ may wrap around the lateral arc surface of the proximal phalanx, and the tendon near the $B'$ end of $B'C'$ may wrap around the lateral arc surface of the middle phalanx. These effects increase local tendon path length and weaken the distal release of $B'$. To keep the model compact, they are combined into
\begin{equation}
\Delta B_3 \approx k_B\,\bigl(r_p-r_p\cos\phi\bigr),
\label{eq:DeltaB3}
\end{equation}
where $k_B$ is an equivalent compensation coefficient for the weakening of distal release caused by lateral wrapping, local deflection, and nonideal contact. In the first order geometric analysis, this term represents higher order path growth that is not explicitly modeled and is not treated as an independent driving mechanism. For a first order two dimensional model, $k_B=0$ can be used; when three dimensional wrapping is included, $k_B$ can be estimated from CAD contact paths or a parameterized wrapping model.

\paragraph{Net distal release and coupled DIP angle}
Combining constrained sliding path shortening, lateral spreading retraction, and additional retraction from lateral wrapping and second order effects, the net distal release of the gliding convergence point $B$ is defined as
\begin{equation}
\Delta B =
\Delta B_1-\Delta B_2-\Delta B_3 .
\label{eq:DeltaB_net}
\end{equation}
Here, $\Delta B_1$ is the dominant release term, whereas $\Delta B_2$ and $\Delta B_3$ are compensation terms in the opposite direction.

Let $r_d$ denote the effective wrapping radius of the terminal tendon at the DIP. The tendon length associated with the net release is approximately converted into the coupled DIP joint angle as
\begin{equation}
\theta_{\text{DIP}}
\approx
\frac{\max(0,\Delta B)}{r_d}
=
\frac{\max\bigl(0,\Delta B_1-\Delta B_2-\Delta B_3\bigr)}{r_d}.
\label{eq:DIP_core}
\end{equation}
The term $\max(0,\Delta B)$ indicates that tendon length allowance for coupled DIP flexion is provided only when lateral band path redistribution produces positive distal release. This equation describes the geometric correspondence between terminal tendon constraint release and coupled DIP angle, rather than active DIP flexion driven by the dorsal terminal tendon.

When the contact point reaches its sliding limit, $\phi$ is fixed at $\phi_{\max}$, and $\Delta B_1$, $\Delta B_2$, and $\Delta B_3$ remain at their limiting values. Further proximal input,
$\Delta F-\Delta F_{\lim}=r_p\bigl(\theta_{\text{PIP}}-\theta_{\text{PIP},\lim}\bigr)$,
is no longer used for wrapping at the PIP and is instead converted \emph{directly} into additional distal displacement of $B$. This work provides a closed form approximation only for the constrained sliding stage, $0\leq\theta_{\text{PIP}}\leq\theta_{\text{PIP},\lim}$.

The resulting PIP--DIP coupling curve is shown in Fig.~\ref{fig19}(J).

\subsection{Joint mechanics of the interossei and lumbricals and the intrinsic-plus posture mechanism}

\begin{figure}[htb]
    \centering
    \includegraphics[width=\columnwidth]{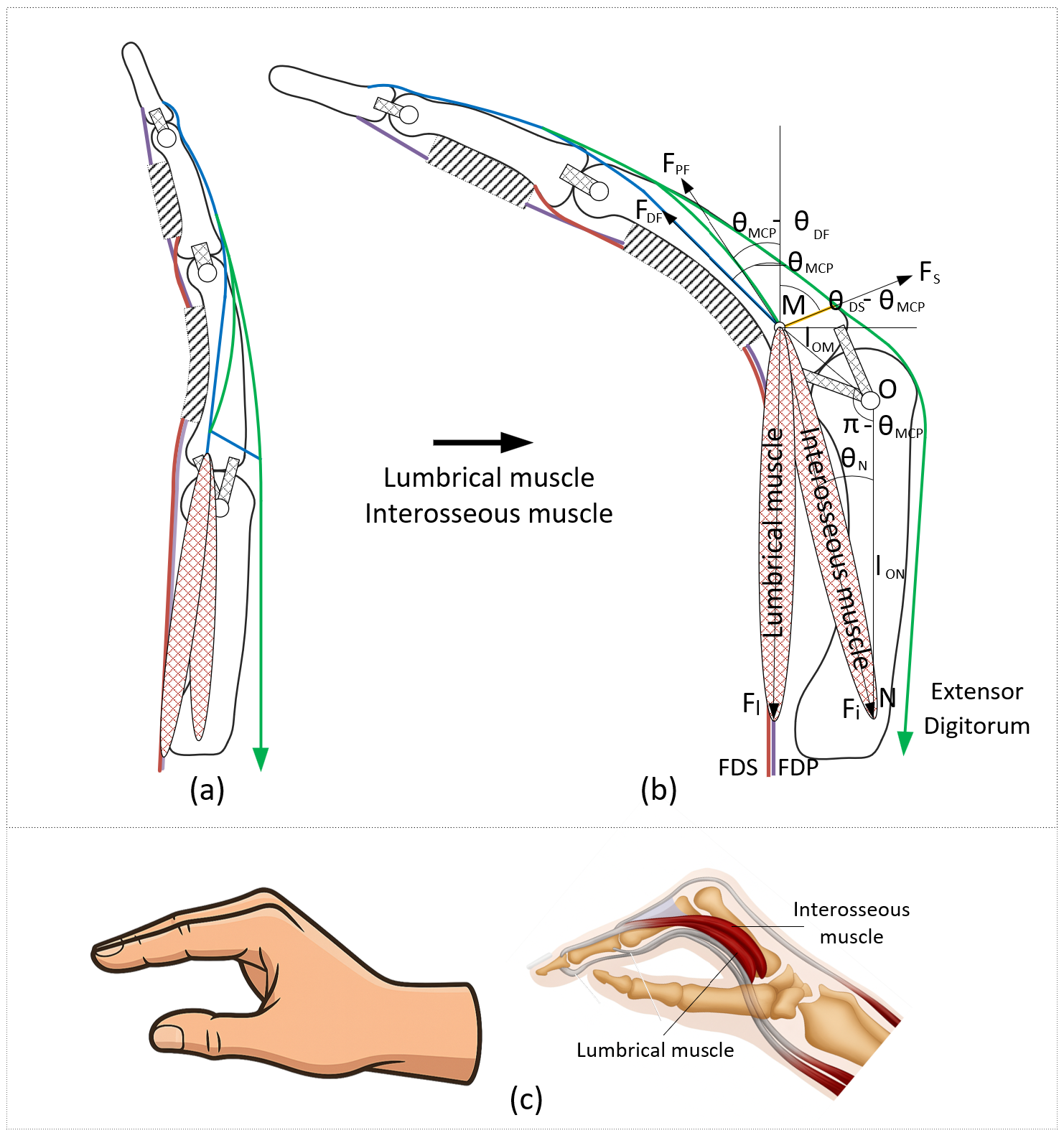}
    \caption{
    Joint mechanics model of the interosseous, lumbrical, and extensor hood system.
    (a,b) Mechanical equilibrium model of the intrinsic muscle extensor hood node at the MCP joint;
    (c) intrinsic-plus posture;
    (d) changes in central slip and lateral band tension with MCP joint angle.
    }
    \label{fig14}
\end{figure}

To establish the mechanical relation among the lumbrical muscle, interosseous muscle, and extensor hood at the MCP node, the force direction of the interosseous muscle is first determined from geometry. The green band denotes the interosseous medial band, which drives PIP return, and the blue band denotes the interosseous lateral band, which drives DIP return. In triangle \(OMN\), \(\angle O=\pi-\theta_{\mathrm{MCP}}\), \(OM=l_{OM}\), and \(ON=l_{ON}\). Planar geometry gives
\begin{equation}
\theta_{FN}
=
\arctan
\left(
\frac{
l_{OM}\sin\theta_{\mathrm{MCP}}
}{
l_{ON}+l_{OM}\cos\theta_{\mathrm{MCP}}
}
\right),
\qquad
\theta_{FN}\in\left(0,\frac{\pi}{2}\right),
\label{eq:thetaFN}
\end{equation}
where \(\theta_{FN}\) is the angle between the interosseous muscle direction and the vertical direction. Because the insertion point of the lumbrical muscle lies on the tendon, its force direction remains approximately vertically downward during joint rotation.

At the connection node among the interosseous muscle, lumbrical muscle, and extensor hood, define
\begin{equation}
\theta_{MD}=\theta_{\mathrm{MCP}}-\theta_{DF},
\qquad
\theta_{DM}=\theta_{DS}-\theta_{\mathrm{MCP}} .
\label{eq:angle-definitions}
\end{equation}
For compact notation, let
\(s_x\equiv\sin\theta_x\) and \(c_x\equiv\cos\theta_x\),
where \(x\in\{MD,\mathrm{MCP},DM,FN\}\).
The planar force equilibrium at the node gives
\begin{equation}
\begin{aligned}
F_{\mathrm{PF}}s_{MD}
+F_{\mathrm{DF}}s_{\mathrm{MCP}}
&=
F_s s_{DM}
+F_i s_{FN},
\\
F_{\mathrm{PF}}c_{MD}
+F_{\mathrm{DF}}c_{\mathrm{MCP}}
+F_s c_{DM}
&=
F_i c_{FN}
+F_l .
\end{aligned}
\label{eq:node-balance}
\end{equation}
Here, \(F_{\mathrm{PF}}\) is the tension in the green band, namely the PIP return band; \(F_{\mathrm{DF}}\) is the tension in the blue band, namely the DIP return band; \(F_s\) is the sagittal band tension; \(F_i\) is the interosseous muscle force; and \(F_l\) is the lumbrical muscle force. The angle \(\theta_{DF}\) denotes the angle between the interosseous lateral band and the interosseous medial band on the extensor hood, and \(\theta_{DS}\) denotes the angle between the sagittal band and the interosseous lateral band.

The following proportional relations are further introduced:
\begin{equation}
F_{\mathrm{PF}}=k_{DP}F_{\mathrm{DF}},
\qquad
F_l=k_{li}F_i ,
\label{eq:ratios}
\end{equation}
where \(k_{DP}>0\) and \(k_{li}>0\) are prescribed proportional coefficients. Treating \(\theta_{DS}\), \(\theta_{DF}\), \(l_{ON}\), and \(l_{OM}\) as known geometric quantities, and substituting Eq.~\eqref{eq:thetaFN} and Eq.~\eqref{eq:ratios} into Eq.~\eqref{eq:node-balance}, yields \(F_{\mathrm{PF}}\) and \(F_{\mathrm{DF}}\), as shown in Fig.~\ref{fig19}(L).

\subsection{Tenodesis mechanism for passive finger flexion driven by wrist extension}

\begin{figure}[htb]
    \centering
    \includegraphics[width=0.48\textwidth]{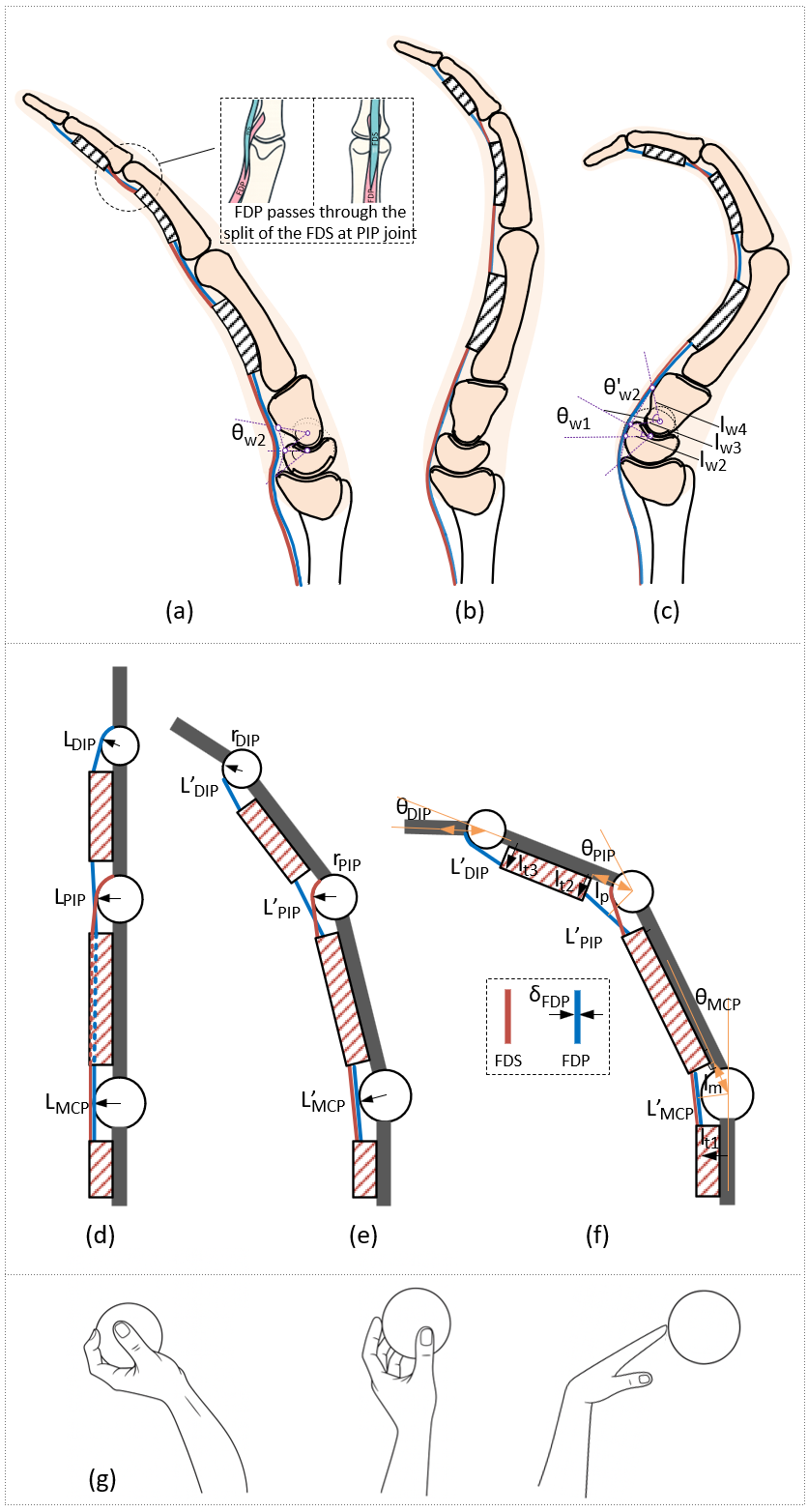}
    \caption{
Tenodesis coupling by which wrist extension induces passive finger flexion.
(a--c) Finger posture during wrist flexion, neutral return, and wrist extension.
(d--f) Schematics of full extension, slight flexion, and pronounced flexion.
(g) Natural coupling between finger joint angles and wrist angle.
(g1--g3) Human ball throwing sequence: preparation with wrist dorsiflexion and finger enclosure, neutral wrist forward swing with partial release, and wrist flexion with full finger extension for ball release.
    }
    \label{fig11}
\end{figure}

As shown in Fig.~\ref{fig11}(a)--(c), wrist--finger coupling is formulated as a tendon length conservation problem for inextensible tendons under geometric guidance. The length injected into the cross wrist tendon path by wrist posture change must be absorbed at the finger side through sheath guided routing and joint wrapping, thereby inducing passive flexion of \(\theta_{\mathrm{MCP}}\), \(\theta_{\mathrm{PIP}}\), and \(\theta_{\mathrm{DIP}}\) without active finger control. The reference posture is defined as the zero state, with wrist angle \(\theta_{w0}\) and fully extended fingers. When the proximal carpal bone rotates by \(\theta_{w1}\) relative to the reference state, the tendon wraps around the proximal wrist with an effective moment arm \(l_{w2}\), producing a length injection linearly related to wrist angle:
\begin{equation}
\Delta L_{ws1}=l_{w2}\theta_{w1}.
\label{eq:w:ws1}
\end{equation}

The distal carpal bone and tendon contact geometry is treated as a triangle formed by the center and two contact points. Its included angle is updated from the reference value \(\theta_{w2}\) to
\begin{equation}
\theta'_{w2}=\theta_{w2}+k_{ws}\theta_{w1},
\label{eq:w:w2_update}
\end{equation}
where \(k_{ws}\) is the proportional coefficient between the angular increments of the distal and proximal carpal bones. For compact notation, define
\begin{equation}
\mathcal{D}_{w}(\alpha)
=
\sqrt{
l_{w3}^{2}+l_{w4}^{2}
-2l_{w3}l_{w4}\cos\alpha
}.
\label{eq:w:Dw}
\end{equation}
By the law of cosines, the equivalent side lengths in the two postures are
\begin{equation}
A=\mathcal{D}_{w}(\theta'_{w2}),
\qquad
B=\mathcal{D}_{w}(\theta_{w2}),
\label{eq:w:AB}
\end{equation}
where \(l_{w3}\) and \(l_{w4}\) are the geometric distances from the center to the two contact points, and \(A\) and \(B\) are the corresponding side lengths under the updated angle \(\theta'_{w2}\) and the reference angle \(\theta_{w2}\), respectively. The tendon length injection contributed by the distal wrist geometry is therefore
\begin{equation}
\Delta L_{ws2}=A-B,
\label{eq:w:ws2}
\end{equation}
and the total wrist side injection is
\begin{equation}
\Delta L_{ws}
=
\Delta L_{ws1}+\Delta L_{ws2}
=
l_{w2}\theta_{w1}
+\mathcal{D}_{w}(\theta'_{w2})
-\mathcal{D}_{w}(\theta_{w2}).
\label{eq:w:ws_total}
\end{equation}

This length injection enters the finger through two parallel paths, the flexor digitorum superficialis (FDS) and flexor digitorum profundus (FDP), and is absorbed at the MCP, PIP, and DIP through different geometric paths. To avoid repeatedly writing long sheath guided routing expressions, define
\begin{equation}
\mathcal{G}(l,t,\theta)
=
2l
-
2\sqrt{t^{2}+l^{2}}\,
\cos\left(
\frac{\theta}{2}
+
\arctan\frac{t}{l}
\right).
\label{eq:w:G}
\end{equation}
This function represents the equivalent length absorption caused by joint flexion angle \(\theta\) when a tendon follows sheath guided routing near a joint.

For the FDP, the tendon undergoes sheath guided geometric absorption when crossing the MCP and PIP, and wraps around an equivalent radius at the DIP. Thus,
\begin{equation}
\begin{aligned}
\Delta L_{\mathrm{MCP-P}}
&=
\mathcal{G}
\left(
l_m,l_{t1},\theta_{\mathrm{MCP}}
\right),\\
\Delta L_{\mathrm{PIP-P}}
&=
\mathcal{G}
\left(
l_p,l_{t2},\theta_{\mathrm{PIP}}
\right),\\
\Delta L_{\mathrm{DIP}}
&=
r_{\mathrm{DIP}}\theta_{\mathrm{DIP}} .
\end{aligned}
\label{eq:w:fdp_segments}
\end{equation}
Here, \(l_m\) and \(l_p\) are the distances from the MCP and PIP centers to the FDP tendon sheath, respectively; \(l_{t1}\) and \(l_{t2}\) are the corresponding sheath offsets; and \(r_{\mathrm{DIP}}\) is the effective wrapping radius of the FDP at the DIP.

For the FDS, because it lies outside the FDP, the finite FDP thickness \(\delta_{\mathrm{FDP}}\) shifts the geometric moment arm of the FDS outward at the MCP. Let
\begin{equation}
l'_m=l_m+\delta_{\mathrm{FDP}},
\qquad
l'_{t1}=l_{t1}+\delta_{\mathrm{FDP}},
\label{eq:w:fds_offset}
\end{equation}
then the length absorption of the FDS at the MCP and PIP is
\begin{equation}
\begin{aligned}
\Delta L_{\mathrm{MCP-S}}
&=
\mathcal{G}
\left(
l'_m,l'_{t1},\theta_{\mathrm{MCP}}
\right),\\
\Delta L_{\mathrm{PIP-S}}
&=
r_{\mathrm{PIP}}\theta_{\mathrm{PIP}} .
\end{aligned}
\label{eq:w:fds_segments}
\end{equation}
where \(r_{\mathrm{PIP}}\) is the effective wrapping radius of the FDS at the PIP.

Under the inextensibility assumption, the wrist side injected length satisfies length balance along both the FDP and FDS paths:
\begin{equation}
\begin{aligned}
\Delta L_{ws}
&=
\Delta L_{\mathrm{MCP-P}}
+
\Delta L_{\mathrm{PIP-P}}
+
\Delta L_{\mathrm{DIP}},\\
\Delta L_{ws}
&=
\Delta L_{\mathrm{MCP-S}}
+
\Delta L_{\mathrm{PIP-S}} .
\end{aligned}
\label{eq:w:balance}
\end{equation}
Eliminating \(\Delta L_{ws}\) gives the closed form expression for the distal wrapping amount at the DIP:
\begin{equation}
\begin{aligned}
\Delta L_{\mathrm{DIP}}
&=
\Delta L_{\mathrm{MCP-S}}
+\Delta L_{\mathrm{PIP-S}}
-\Delta L_{\mathrm{MCP-P}}
-\Delta L_{\mathrm{PIP-P}}                                      \\
&=
\left(\Delta L_{\mathrm{MCP-S}}
-\Delta L_{\mathrm{MCP-P}}\right)                               \\
&\quad+
\left(\Delta L_{\mathrm{PIP-S}}
-\Delta L_{\mathrm{PIP-P}}\right).
\end{aligned}
\label{eq:w:Delta_ldip}
\end{equation}

At the MCP, because the FDS is more superficial and is affected by the FDP thickness \(\delta_{\mathrm{FDP}}\), its effective moment arm is slightly larger than that of the FDP, so
\(\Delta L_{\mathrm{MCP-S}}-\Delta L_{\mathrm{MCP-P}}>0\).
At the PIP, the effective moment arm of the split FDS is usually slightly smaller than that of the FDP, so
\(\Delta L_{\mathrm{PIP-S}}-\Delta L_{\mathrm{PIP-P}}<0\).
In this geometric arrangement, the two difference terms largely cancel each other, leaving a relatively small effective excursion for the DIP. This explains the classical structure in which the FDS splits at the PIP and allows the FDP to pass through: the geometric absorption differences at the proximal and middle phalanges are partially compensated, thereby limiting excessive passive flexion of the DIP.

In the numerical implementation, \(\Delta L_{ws}(\theta_{w1})\) is first computed from Eqs.~\eqref{eq:w:ws1}--\eqref{eq:w:ws_total}. Eq.~\eqref{eq:w:balance} is then solved together with the joint ratio relation
\begin{equation}
\theta_{\mathrm{PIP}}=k_{PM}\theta_{\mathrm{MCP}}
\label{eq:w:pip_mcp_ratio}
\end{equation}
where \(k_{PM}\) is the angular proportionality coefficient between the MCP and PIP, obtained from the joint impedance ratio. The values of \(\theta_{\mathrm{MCP}}\), \(\theta_{\mathrm{PIP}}\), and \(\theta_{\mathrm{DIP}}\) are then solved, and \((1+k_{ws})\theta_{w1}\) is used as the effective wrist extension angle to plot the wrist--finger passive coupling curve.

Figure~\ref{fig19}(K) shows the relation between wrist posture and passive finger joint responses, indicating that wrist--finger coupling can transform wrist posture changes into finger pre shaping input. Figure~\ref{fig11}(g) further illustrates the role of this mechanism in grasp and release motion: wrist extension induces finger flexion through cross wrist tendon paths and stabilizes the object, whereas wrist flexion promotes finger opening and makes the fingertip pad normal force direction more favorable for throwing during release.

% ===========================================================
\section{Conclusion}
% ===========================================================

MCR-Bionic establishes a 1:1 wrist and hand hardware platform in which a two row eight bone wrist, the wrist ligament network, more than 103 whole hand ligament/soft tissue limit structures, anatomical FDS/FDP routing, volar plates and collateral ligaments, tendon sheaths, the dorsal extensor hood, intrinsic muscle pathways, and local in-hand artificial muscle actuation are integrated within a single mechanical body. The importance of this integration is not the number of reconstructed parts alone, but the use of ligament and soft tissue to define compliant joint boundaries, small off axis motion, and contact adaptability close to those of the human hand.

On this hardware basis, human like dexterous manipulation is reformulated as a problem shared by structure and control. Wrist--finger tenodesis provides grasp pre shaping, the extensor hood establishes PIP--DIP distal coordination, and intrinsic-plus related pathways regulate MCP posture, distal stability, and fingertip force paths after grasp formation. Contact rich tasks such as coin flipping, dorsal transfer, pen manipulation, and Rubik's cube pushing show that fine in-hand manipulation depends not only on additional degrees of freedom or more complex control, but also on compliant physical boundaries formed by soft tissue limits, extensor hood differentiation, and muscle pathways.

Thus, the aim of anatomical biomimetics is not to copy all biological details, but to identify which structures change the relations among input, motion, contact, and stability in the robotic body. For human like dexterous hands, the key question is not how human like the hand should appear, but which structures, once removed, change the control problem and the manipulable states themselves. MCR-Bionic turns this question into a testable robotics platform, suggesting that some structures of the human hand are not morphological inheritance, but part of dexterous manipulation itself.

\section*{Acknowledgments}

\textbf{Use of AI assisted tools:} AI-assisted tools were used for language polishing, translation assistance, consistency checking, and background removal in selected figure images. They were not used to generate experimental data, analyses, scientific conclusions, or quantitative results. All assisted text and image edits were reviewed by the authors, who take full responsibility for the final manuscript.

% ==========================================================

\end{CJK*}
\end{document}